\documentclass{article}

    \usepackage[preprint]{tackling_climate_workshop_style}

\usepackage[utf8]{inputenc} 
\usepackage[T1]{fontenc}    
\usepackage{hyperref}       
\usepackage{url}            
\usepackage{booktabs}       
\usepackage{multirow}
\usepackage{amsfonts}       
\usepackage{amsmath,amsthm,amssymb}
\usepackage{mathtools}
\usepackage{enumitem}
\usepackage{nicefrac}       
\usepackage{microtype}      
\usepackage{tikz}
\usepackage{comment}
\usepackage{eurosym}

\usepackage{natbib}
\newcommand{\minimize}[1]{\underset{{#1}}{\text{minimize}}}

\newcommand{\st}{\text{subject to}}

\DeclareMathAlphabet{\mymathbb}{U}{BOONDOX-ds}{m}{n}
\DeclareMathSymbol{\shortminus}{\mathbin}{AMSa}{"39}
\def\shortplus{
\begin{tikzpicture}
    \node[draw=none,scale=0.45,inner sep=0] at (0,0) {$\boldsymbol{+}$};
\end{tikzpicture}
}

\usepackage{xcolor}
\definecolor{niceblue}{rgb}{0, 0.5, 1.0}

\hypersetup{
    colorlinks=true,
    linkcolor=niceblue,
    filecolor=magenta,      
    urlcolor=niceblue,
    citecolor=niceblue,
    }

\title{Emission-Aware Optimization of Gas Networks: Input-Convex Neural Network Approach}

\author{%
  Vladimir Dvorkin \\
  MIT Energy Initiative \\
  Massachusetts Institute of Technology\\
  \texttt{dvorkin@mit.edu} \\
  \And
  Samuel Chevalier, Spyros Chatzivasileiadis \\
  Department of Wind and Energy Systems\\
  Technical University of Denmark \\
  \texttt{\{schev,spchatz\}@elektro.dtu.dk} \\
}

\begin{document}

\maketitle

\begin{abstract}
    Gas network planning optimization under emission constraints prioritizes gas supply with the least CO$_2$ intensity. As this problem includes complex physical laws of gas flow, standard optimization solvers cannot guarantee convergence to a feasible solution. To address this issue, we develop an input-convex neural network (ICNN) aided optimization routine which incorporates a set of trained ICNNs approximating the gas flow equations with high precision. Numerical tests on the Belgium gas network demonstrate that the ICNN-aided optimization dominates non-convex and relaxation-based solvers, with larger optimality gains pertaining to stricter emission targets. Moreover, whenever the non-convex solver fails, the ICNN-aided optimization provides a feasible solution to network planning.
\end{abstract}

\section{Introduction}

Energy system operators worldwide introduce carbon reduction measures to control the CO$_2$ content of energy supply~\cite{liu2021country}. Such measures include soft monetary penalties (e.g., carbon tax) or hard emission constraints (e.g., carbon cap or renewable portfolio standard) within operational planning routines. When operation planning is insufficient, more expensive yet effective long-term planning optimizes network design in order to accommodate more environment-friendly supply.

In gas networks, which connect spatially distributed supply and demand hubs, carbon reduction measures can be used to prioritize pipeline-quality gas from renewable suppliers, such as biogas produced from organic matter, syngas produced by steam reforming, or hydrogen produced from electrolysis at large offshore wind sites, like the proposed North Sea hub~\cite{NSWPH}. However, solving such planning optimization problems with emission targets is challenging due to complex gas flow physics.

\textbf{Contributions.} To address gas network planning under emission targets, we devise a new optimization method that substitutes the non-convex Weymouth equation of gas flows with a composition of trained input-convex and input-concave neural networks (ICNNs). Together, they explain the dependency of gas flows on nodal pressures. We embed trained ICNNs into planning optimization problems which are then solved using standard mixed-integer solvers. Tests on the Belgium gas network demonstrate the improvement of our methods over standard solvers, especially under strict emission targets. 

\subsection{Related work}

\textbf{Gas network optimization.} Designing optimization methods to aid operation planning dates back to at least 1979 \cite{o1979mathematical}. Since then, solvers based on mixed-integer  \cite{wilson1988steady}, piece-wise linear \cite{de2000gas}, quadratic \cite{singh2019natural,singh2020natural} and semi-definite \cite{ojha2017solving} programming have been introduced. The CO$_{2}$ footprint of integrated gas and electricity networks has been addressed by integrating renewables \cite{ordoudis2019integrated,ratha2020affine,roald2020uncertainty,dvorkin2021stochastic,dvorkin2022multi} or directly incorporating carbon reduction measures in operational  \cite{piperagkas2011stochastic,cheng2019low} and long-term expansion planning problems \cite{degleris2021emissionsaware,qiu2014low,cheng2018planning}. We refer to \cite{conejo2020operations} for a comprehensive literature review. 

\textbf{Neural networks to aid optimization.} Using the mixed-integer neural network (NN) reformulation \cite{tjeng2017evaluating,xiao2018training,grimstad2019relu}, NNs can be used for approximating complex input-output dependencies within optimization, e.g., in power systems problems \cite{murzakhanov2020neural,donon2020neural,hu2020physics,kody2022modeling}. The reformulation represents the activation of each ReLU function using linear and binary constraints parameterized by NN weights and biases, which can be computationally challenging at scale \cite{grimstad2019relu}. Here, we explore an alternative functional approximation that relies on \textit{input-convex} NNs, which constrain network weights to ensure the output is a convex function of inputs \cite{amos2017input}. Since trained ICNN mappings can be recast as linear optimization problems \cite{amos2017input,duchesne2021supervised}, we leverage them to convert non-convex optimization problems into bilevel optimization problems which are linear in both their upper- and lower-levels \cite{pozo2017basic}.

\section{Emission-aware gas network planning problems}
\textbf{Operational planning problem.} A gas network includes $n$ nodes, representing injections, extractions or network junctions, and $\ell$ edges, representing pipelines. The operational planning problem identifies the least-cost supply allocation $\vartheta\in\mathbb{R}^{n}$ that satisfies nodal gas demands $\delta\in\mathbb{R}^{n}$, while ensuring that nodal pressures $\pi\in\mathbb{R}^{n}$ and gas flows $\varphi\in\mathbb{R}^{\ell}$ remain within technical limits. This problem is solved using the following optimization formulation~\cite{dvorkin2021stochastic}:
\begin{subequations}\label{OGF}
\begin{align}
    \minimize{{\varphi,\vartheta,\pi\in\mathcal{F}}}\quad&c^{\top}\vartheta\label{OGF:obj}\\
    \st\quad&A\varphi=\vartheta-\delta,\label{OGF:balance}\\
    &\varphi\circ|\varphi|=\text{diag}[\omega]A^{\top}\pi,\label{OGF:weymouth}
\end{align}
\end{subequations}
which minimizes linear gas supply costs subject to technical constraints. Using graph admittance matrix $A\in\mathbb{R}^{n\times \ell}$, equation \eqref{OGF:balance} ensures the conservation of gas mass. Given the fixed friction coefficients $\omega\in\mathbb{R}^{\ell}$, the steady-state Weymouth equation \eqref{OGF:weymouth} enforces the non-convex dependency of gas flows on pressure variables. Finally, a convex set $\mathcal{F}$ is used to respect the technical limits on gas mass and pressures. Note that vector $\pi$ contains squared nodal pressures to reduce non-linearities in \eqref{OGF:weymouth} \cite{de2000gas}. We do not model compressors, which can be incorporated with fixed \cite{singh2019natural,singh2020natural} or varying \cite{de2000gas,dvorkin2021stochastic} compression rates without significant impacts on computational costs.  

Although cost function \eqref{OGF:obj} typically includes only marginal production costs, it can also internalize an emission (carbon) tax to penalize gas producers with higher environmental impact. Alternatively, emissions can be regulated by carbon cap constraints on the total emission level. Although the equivalence of carbon tax and carbon cap can be shown through the Karush–Kuhn–Tucker conditions of \eqref{OGF} \cite{brown2021decreasing}, the carbon cap is preferred due to non-convexities in \eqref{OGF:weymouth}. Indeed, the same emission goal may not be achieved under the carbon tax, since local search algorithms may fail to minimize the penalty term globally; meanwhile, the carbon cap is introduced through the hard constraint, i.e.,
\begin{align}\label{eq:emission_cap}
    e^{\top}\vartheta\leqslant \overline{e},
\end{align}
with vector $e\in\mathbb{R}^{n}$ of carbon intensities and carbon cap $\overline{e}$, which must be satisfied at all times. 

\textbf{Long-term planning problem.}\label{para_plan} Since a carbon cap may significantly affect the operating cost in \eqref{OGF:obj}, the long-term planning problem optimizes the network design to enable more economical satisfaction of the emission constraint \eqref{eq:emission_cap}. This problem is especially relevant for the design of future hydrogen gas transport networks which governments are actively considering~\cite{khan2021techno}. Let the diameter $d\in\mathbb{R}^{\ell}$ of gas pipelines be the design variable. Since pipeline friction is often modeled as being linearly proportional to diameter~\cite{Sundar:2019}, a constant $\hat{\omega}_i$ can be used to relate friction and diameter via $\omega_{i}=\hat{\omega}_{i}d_{i}$. The diameter enters the operational problem \eqref{OGF} through the Weymouth equation \eqref{OGF:weymouth} as
\begin{align}\label{expantion:Weymouth}
 & \text{diag}[d]^{-1}\varphi\circ|\varphi|=\text{diag}[\hat{\omega}]A^{\top}\pi,
\end{align}
where the right-hand side has no explicit dependence on diameter. By defining a vector $\lambda\in\mathbb{R}^{\ell}$ of expansion costs, we obtain a long-term planning optimization from problem \eqref{OGF} by augmenting the total cost of expansion $\lambda^{\top}d$ to \eqref{OGF:obj} and substituting equation \eqref{OGF:weymouth} with its dynamic counterpart in \eqref{expantion:Weymouth}. 

\section{Input-convex neural network approach to emission-aware planning} 

Addressing the non-convex equation \eqref{OGF:weymouth}, we observe that its left-hand side $f(\varphi_{l})=\varphi_{l}|\varphi_{l}|$ is convex for $\varphi_{l}\geqslant0$ and concave for $\varphi_{l}\leqslant0$. Hence, $f(\varphi_{l})$ can be approximated with a sum $f(\varphi_{l})\approx\Phi_{\shortplus}(\varphi_{l})+\Phi_{\shortminus}(\varphi_{l})$ of one input-convex $\Phi_{\shortplus}(\varphi_{l})$ and one input-concave $\Phi_{\shortminus}(\varphi_{l})$ neural network. 
We use the following $k-$layer architectures under ReLU activation functions of hidden neurons: 
\begin{align*}
    \begin{array}{l}
    \Phi_{\shortplus}(\varphi_{l}) \colon\quad z^{1}_{\shortplus} = \text{max}\left(0,W^{0}_{\shortplus}\varphi_{l}+b^{0}_{\shortplus}\right), 
    \quad
    z^{i+1}_{\shortplus} = \text{max}\left(0,W^{i}_{\shortplus}z^{i}_{\shortplus}+b^{i}_{\shortplus}\right), \forall i=1,\dots,k-1,\\[1.0ex]
    \Phi_{\shortminus}(\varphi_{l}) \colon\quad z^{1}_{\shortminus} = \text{max}\left(0,W^{0}_{\shortminus}\varphi_{l}+b^{0}_{\shortminus}\right), 
    \quad
    z^{i+1}_{\shortminus} = \text{max}\left(0,W^{i}_{\shortminus}z^{i}_{\shortminus}+b^{i}_{\shortminus}\right), \forall i=1,\dots,k-1,
    \end{array}
\end{align*}
with a scalar input $\varphi_{l}$, scalar output $z_{k}$, and weights and biases $W$ and $b$, respectively. In $\Phi_{\shortplus}(\varphi_{l})$, the weights $W^{i}_{\shortplus},\forall i=1,\dots,k-1$ are non-negative to render the output a convex function of the input. In $\Phi_{\shortminus}(\varphi_{l})$, the weights $W^{i}_{\shortminus}$ are also non-negative for $i=1,\dots k-2$, but they are non-positive  for $i=k-1$ to render the output a concave function of the input. With such architectures, we have a piece-wise functional approximation $f(\varphi_{l})\rightarrow z_{\shortplus}^{k} + z_{\shortminus}^{k}$. From \cite[Appendix B]{amos2017input}, we can retrieve the output of the trained ICNNs from the input  by solving a linear program, e.g.,
\begin{subequations}\label{prog:icnn_ref_base}
\begin{align}
     \minimize{z^{1}_{\shortplus},\dots,z^{k}_{\shortplus}}\quad& z^{k}_{\shortplus}\\
     \st\quad&z^{1}_{\shortplus}\geqslant W^{0}_{\shortplus}\varphi_{l}+b^{0}_{\shortplus}, \quad z^{i+1}_{\shortplus}\geqslant  W^{i}_{\shortplus}z^{i}_{\shortplus}+b^{i}_{\shortplus}, \quad z^{i}_{\shortplus}\geqslant \mymathbb{0}, \quad \forall i=1,\dots,k-1,
\end{align}
\end{subequations}
for the $\Phi_{\shortplus}(\varphi_{l})$ architecture, and it takes a similar form for the $\Phi_{\shortminus}(\varphi_{l})$ architecture. Thus, to approximate the Weymouth equation, we need to embed two linear programs (one convex and one concave) for each pipeline. The computational burden, however, will depend on the number of hidden layers and neurons. To reduce the burden, we note that for $\varphi_{l}\geqslant0$, solution $z^{k}_{\shortplus}$ is an {\it outer approximation} of the trained ICNN output, and the number of approximating hyperplanes $2^{p}$ is the number of unique combinations of $p$ hidden neurons. For small -- yet sufficient to represent a convex function -- architectures, we can screen approximating hyperplanes and leave only a set $\mathbb{H}_{\shortplus}$ of {\it supporting} hyperplanes, for which there exists an input $\varphi_{l}$ which makes such hyperplanes active (binding). Such hyperplane parameters are obtained from the trained ICNN as
\begin{align*}
&\textstyle\prod_{r=k}^{0}(s_{j}^{r}\circ W_{\shortplus}^{r})\varphi_{l}+\textstyle\sum_{i=0}^{k}\prod_{r=k}^{i}(s_{j}^{r}\circ W_{\shortplus}^{r})b^{i} = w_{\shortplus}^{j}\varphi_{l} + v_{\shortplus}^{j},\quad\forall j \in\mathbb{H}_{\shortplus},
\end{align*}
with slope $w_{\shortplus}^{j}$ and intercept $v_{\shortplus}^{j}$. Vector $s_{j}\in\mathbb{R}^{p}$ collects a unique combination of ReLU activations (1 if active, and 0 if otherwise) of hyperplane $j$, and $s_{j}^{r}$ is a subset of $s_{j}$ with hidden neurons of layer $r$. Similarly, we obtain the active hyperplanes for the outer approximation of the concave part of $f(\varphi_{l})$. 

We now put forth the bilevel operational planning optimization which embeds the trained ICNNs:
\begin{subequations}\label{bilevel}
\begin{align}
    \minimize{{\varphi,\vartheta,\pi\in\mathcal{F}}}\quad&c^{\top}\vartheta\\
    \st\quad&\text{Constraints}\;\eqref{OGF:balance},\eqref{eq:emission_cap},\quad t_{\shortplus} + t_{\shortminus}=\text{diag}[\omega]A^{\top}\pi,\\
        &\begin{matrix*}[l]
         t_{\shortplus}^{l} \in &  \text{minimize}_{\;t_{\shortplus}^{l}} & t_{\shortplus}^{l}, \quad \st\;  w_{\shortplus}^{i}\varphi_{l}+ v_{\shortplus}^{i}\leqslant t_{\shortplus}^{l},\quad\forall i\in\mathbb{H}_{\shortplus}, \forall l\in1,\dots,e
    \end{matrix*}\label{Bilevel:LL_convex}\\
    &\begin{matrix*}[l]
         t_{\shortminus}^{l} \in &  \text{maximize}_{\;t_{\shortminus}^{l}} & t_{\shortminus}^{l},\quad \st\; w_{\shortminus}^{i}\varphi_{l}+v_{\shortminus}^{i}\geqslant t_{\shortminus}^{l},\quad\forall i\in\mathbb{H}_{\shortminus}, \forall l\in1,\dots,e
    \end{matrix*}\label{Bilevel:LL_concave}
\end{align}
\end{subequations}
where \eqref{Bilevel:LL_convex} and \eqref{Bilevel:LL_concave} are lower-level optimization problems, each including a single auxiliary variable $t^{l}$ which returns the ICNN output. Indeed, problem \eqref{Bilevel:LL_convex} is a light-weighted version of \eqref{prog:icnn_ref_base} producing the identical approximation result. Appendix \ref{app:ICNN_KKT_reformulation} provides a tractable mixed-integer reformulation of \eqref{bilevel} using Karush–Kuhn–Tucker (KKT) conditions of \eqref{Bilevel:LL_convex} and \eqref{Bilevel:LL_concave}. Then, in Appendix \ref{app:Weymputh_explantion_details}, we show that the dynamic Weymouth equation \eqref{expantion:Weymouth} can also be approximated by an ICNN.

\section{Numerical tests on the Belgium gas network}

To demonstrate emission-aware planning, we use a modified Belgium system from \cite{de2000gas}, with a meshed topology, tighter pressure bounds, and more distributed gas supply and demand hubs. Using this system, we compare three methods to solve operation planning: 1) an interior point solver \texttt{IPOPT}~\cite{wachter2006implementation}, 2) a mixed-integer quadratic programming (MIQP) relaxation, detailed in Appendix \ref{app:MIQP_relaxation}, and 3) the proposed ICNN-aided optimization. The last two are solved with mixed-integer Gurobi solver~\cite{gurobi}. The long-term planing is solved by the 1\textsuperscript{st} and 3\textsuperscript{rd} methods only, as no convex relaxation of equation \eqref{expantion:Weymouth} is known. The CPU time for all methods does not exceed several minutes. The NN architectures include 1 hidden layer with up to 15 neurons, which was sufficient to approximate convex and concave parts of the Weymouth equation. Test data, details on the training procedure, and codes to replicate our results are available at \url{https://doi.org/10.5281/zenodo.7089330}

The CO$_{2}$ intensity of the gas supply in the test system varies between 0.6 and 2.7 kg/m$^3$, and solving the operational planning problem \eqref{OGF} without emission constraint \eqref{eq:emission_cap} results in up to 125.9 kT of emitted CO$_{2}$ with the \texttt{IPOPT} solver. To limit emissions, we select one moderate emission cap of 100 kT and one extreme cap of 48.9 kT, below which no method returns a feasible solution. 

The solutions for operation planning are collected in Table \ref{tab:operation}. As emission cap reduces, the \texttt{IPOPT} solver becomes more sensitive to initialization and fails to provide a feasible solution with probability up to 39.0\%. Although the termination status of the MIQP relaxation is always optimal, it may not be feasible with respect to the original, non-relaxed Weymouth equation; using it as a warm start for \texttt{IPOPT}, however, we retrieve a feasible point, which is competitive with the best performance of randomly initialized \texttt{IPOPT} solver. With either moderate or no emission cap, the proposed ICNN-aided optimization improves on the MIQP solution and consistently returns the best solution found with \texttt{IPOPT}. In the most constrained case, with $\overline{e}=48.9$  kT, the ICNN-aided optimization solution provides the least-cost operation cost, thus dominating both \texttt{IPOPT} and MIQP solutions.

Table \ref{tab:coopt} provides the summary of long-term planning cost, which includes both operating cost and adjusted (to a single, peak hour) expansion cost. While the \texttt{IPOPT} solver exhibits a large variance and fails to produce any solution with probability up to 41.4\%, the ICNN-aided optimization always returns the best solution discovered with random \texttt{IPOPT} initializations. For the worst case \texttt{IPOPT} outcomes, the ICNN-aided solution yields 3.2\%--5.9\% cost savings, as it requires less pipeline expansion; e.g.,  for $\overline{e}=48.9$ kT, it expands pipelines by 117mm less on average across the network.

\begin{table}[h]
\caption{Cost summary of the emission-aware operation planning (\euro1,000). 
}
\label{tab:operation}
\begin{tabular}{cc@{\hspace{1.25\tabcolsep}}c@{\hspace{1.25\tabcolsep}}c@{\hspace{1.25\tabcolsep}}cc@{\hspace{1.25\tabcolsep}}cc@{\hspace{1.25\tabcolsep}}c}
\toprule
\multirow{2}{*}{\begin{tabular}[c]{@{}c@{}}Emission \\ cap, kT\end{tabular}} & \multicolumn{4}{c}{1,000 random \texttt{IPOPT} initializations}                                       & \multicolumn{2}{c}{MIQP relaxation} & \multicolumn{2}{c}{ICNN-aided solution} \\
\cmidrule(lr){2-5}
\cmidrule(lr){6-7}
\cmidrule(lr){8-9}
                     & min    & mean   & max    & \begin{tabular}[c]{@{}c@{}}prob. of\\ failure\end{tabular} & optimal             & \begin{tabular}[c]{@{}c@{}}warm start\\ for \texttt{IPOPT}\end{tabular}            & optimal        & \begin{tabular}[c]{@{}c@{}}warm start\\ for \texttt{IPOPT}\end{tabular}        \\
\midrule
$\infty$                 & 1,923.3 & 1,927.2 & 1,929.2 & 16.6\%                                                       & 1,540.8           & 1,929.2           & 1,932.3      & 1,923.3       \\
100                 & 2,225.1 & 2,235.1 & 2,256.2 & 16.0\%                                                       & 2,137.2           & 2,225.1           & 2,241.3      & 2,225.1       \\
48.9                  & 4,344.6 & 4,344.6 & 4,344.6 & 39.0\%                                                       & 4,200.8           & 4,344.6           & 4,290.1        & 4,291.2      \\
\bottomrule
\end{tabular}
\end{table}

\begin{table}[h]
\centering
\caption{Cost summary of the emission-aware long-term planning (\euro1,000).
}\label{tab:coopt}
\begin{tabular}{cc@{\hspace{1.5\tabcolsep}}c@{\hspace{1.5\tabcolsep}}c@{\hspace{1.5\tabcolsep}}cc@{\hspace{1.5\tabcolsep}}c}
\toprule
\multirow{2}{*}{\begin{tabular}[c]{@{}c@{}}Emission \\ cap, kT\end{tabular}} & \multicolumn{4}{c}{1,000 random \texttt{IPOPT} initializations}                                       & \multicolumn{2}{c}{ICNN-aided solution} \\
\cmidrule(lr){2-5}
\cmidrule(lr){6-7}
                     & min    & mean   & max    & \begin{tabular}[c]{@{}c@{}}prob. of\\ failure\end{tabular} & optimal        & \begin{tabular}[c]{@{}c@{}}warm start\\ for \texttt{IPOPT}\end{tabular}        \\
\midrule
$\infty$                  & 2,671.7 & 2,701.8 & 2,829.5 & 28.6\%                                                                & 2,666.4      & 2,671.6       \\
100                 & 3,057.8 & 3,090.2 & 3,191.9 & 30.3\%                                                                 & 3,056.6      & 3,057.8       \\
48.9                  & 5,079.1 & 5,138.7 & 5,247.9 & 41.4\%                                                                  & 5,079.9        & 5,079.1      \\
\bottomrule
\end{tabular}
\end{table}

\section{Conclusion}
We developed a new method for operation and long-term planning of gas networks under emission constraints, based on embedding input-convex and input-concave neural networks into planning optimization. We demonstrated empirical evidence that our method is robust even to strictest emission targets, for which the non-convex and relaxation-based solvers often fail to produce a feasible solution. A cost-saving potential of our method is up to 1.2\% in operational and 5.9\% in long-term planning.  

\section*{Acknowledgements}
Vladimir Dvorkin is supported by the MSCA-COFUND Postdoctoral Program, Grant Agreement No. 101034297 -- project Learning ORDER. Samuel Chevalier  is supported by the HORIZON-MSCA-2021 Postdoctoral Fellowship Program, Grant Agreement No. 101066991 -- project TRUST-ML. Spyros Chatzivasileiadis is supported by the ERC Starting Grant, Grant Agreement No. 949899 -- project VeriPhIED.

\newpage
{\small
\bibliography{references.bib}}

\appendix 
\section{Reformulation of ICNN-aided optimization via KKT conditions}\label{app:ICNN_KKT_reformulation}
We consider the lower level problem from \eqref{Bilevel:LL_convex} associated with the input convex NN (the concave case is dealt with similarly) for a single line $l$:
\begin{subequations}\label{eq: conv_ll}
\begin{align}
     \minimize{t_{\shortplus}}\quad& t_{\shortplus}\\
     \st\quad& w_{\shortplus}^{i}\varphi+v_{\shortplus}^{i}\leqslant t_{\shortplus} : \mu_{\shortplus}^{i}\in\mathbb{R}^{E},\quad\forall i\in\mathbb{H}_{\shortplus},\label{eq: conv_ll_ineq}
\end{align}
\end{subequations}
where $\mu_{\shortplus}^{i}$ is the Lagrange multiplier associated with the $i^{\rm th}$ inequality constraint. The Lagrangian function~\cite{boyd2004convex} associated with this linear program is given by
\begin{subequations}
\begin{align}
    L(t_{\shortplus},\mu_{\shortplus}) =  t_{\shortplus} + \sum_{i\in\mathbb{H}_{\shortplus}}\mu_{\shortplus}^{i\top}(w_{\shortplus}^{i}\varphi+v_{\shortplus}^{i}- t_{\shortplus}).
\end{align}
\end{subequations}
The KKT conditions associated with the linear program \eqref{eq: conv_ll} may now be derived. Primal and dual feasibility can be directly stated:
\begin{equation}
\begin{aligned}
    \begin{array}{lll}
    \text{primal feasibility:}  & w_{\shortplus}^{i}\varphi+v_{\shortplus}^{i}\leqslant t_{\shortplus},& \forall i\in\mathbb{H}_{\shortplus} 
    \\
    \text{dual feasibility:}  & \mu_{\shortplus}^{i} \geqslant 0, & \forall i\in\mathbb{H}_{\shortplus}.
    \end{array}
\end{aligned}
\end{equation}
The stationarity condition can be computed by taking the derivative of the Lagrangian (with respect to the primary variable) and setting it equal to 0:
\begin{align}
\frac{\partial}{\partial t_{\shortplus}}L(t_{\shortplus},\mu_{\shortplus})=1-\sum_{i\in\mathbb{H}_{\shortplus} }\mu_{\shortplus} \equiv 0.
\end{align}
Therefore, stationarity and complementary slackness are given as
\begin{equation}
\begin{aligned}
    \begin{array}{lll}
    \text{stationarity condition:}  & \sum_{i\in\mathbb{H}_{\shortplus}}\mu_{\shortplus}^{i} = 1, & 
    \\
    \text{complementary slackness:}  & \mu_{\shortplus}^{i}\cdot(w_{\shortplus}^{i}\varphi+v_{\shortplus}^{i}- t_{\shortplus})=0, & \forall i\in\mathbb{H}_{\shortplus}.
    \end{array}
\end{aligned}
\end{equation}
Since only one inequality constraint in \eqref{eq: conv_ll_ineq} can be active, the dual variables are implicitly constrained to be binary: $\mu_{\shortplus}^{i}\in\{0,1\},\, \forall i \in\mathbb{H}_{\shortplus}$, but only one may take a nonzero value. Since the dual variables are constrained to be binary, the quadratic complimentary slackness constraints can be effectively linearized using Big-M:
\begin{align}
\mu_{\shortplus}^{i}(w_{\shortplus}^{i}\varphi+v_{\shortplus}^{i}- t_{\shortplus})=0
\quad \Leftrightarrow \quad (\mu_{\shortplus}^{i} - 1)M \leqslant w_{\shortplus}^{i}\varphi+\mymathbb{1}v_{\shortplus}^{i} - t_{\shortplus}
\end{align}
where no upper bound is needed, since $w_{\shortplus}^{i}\varphi+\mymathbb{1}v_{\shortplus}^{i} - t_{\shortplus} \leqslant 0$ is implied by primal feasibility. The final KKT reformulation of the lower-level problem is:
\begin{equation}
\begin{aligned}
    \begin{array}{ll}
    w_{\shortplus}^{i}\varphi+v_{\shortplus}^{i}\leqslant t_{\shortplus},& \forall i\in\mathbb{H}_{\shortplus} 
    \\
    (\mu_{\shortplus}^{i}-1)M \leqslant w_{\shortplus}^{i}\varphi+v_{\shortplus}^{i}- t_{\shortplus} \leqslant 0, & \forall i\in\mathbb{H}_{\shortplus}
    \\
    \sum_{i\in\mathbb{H}_{\shortplus}}\mu_{\shortplus}^{i} = 1, \quad \mu_{\shortplus}^{i}\in\{0,1\}, & \forall i\in\mathbb{H}_{\shortplus}.
    \end{array}
\end{aligned}
\end{equation}
Similarly, the KKTs of the concave lower-level problem are: 
\begin{equation}
\begin{aligned}
    \begin{array}{ll}
    w_{\shortminus}^{i}\varphi+v_{\shortminus}^{i}\geqslant t_{\shortminus},& \forall i\in\mathbb{H}_{\shortminus} 
    \\
    0\leqslant w_{\shortminus}^{i}\varphi+v_{\shortminus}^{i}- t_{\shortminus} \leqslant (1-\mu_{\shortminus}^{i})M, & \forall i\in\mathbb{H}_{\shortminus}
    \\
    \sum_{i\in\mathbb{H}_{\shortminus}}\mu_{\shortminus}^{i} = 1, \quad \mu_{\shortminus}^{i}\in\{0,1\}, & \forall i\in\mathbb{H}_{\shortminus}.
    \end{array}
\end{aligned}
\end{equation}
Both of these formulations are additionally applicable for neural networks which map multiple inputs (rather than just a single input $\varphi$) to scalar outputs.

\begin{figure}[b!]
\centering
\includegraphics[width=0.7\columnwidth]{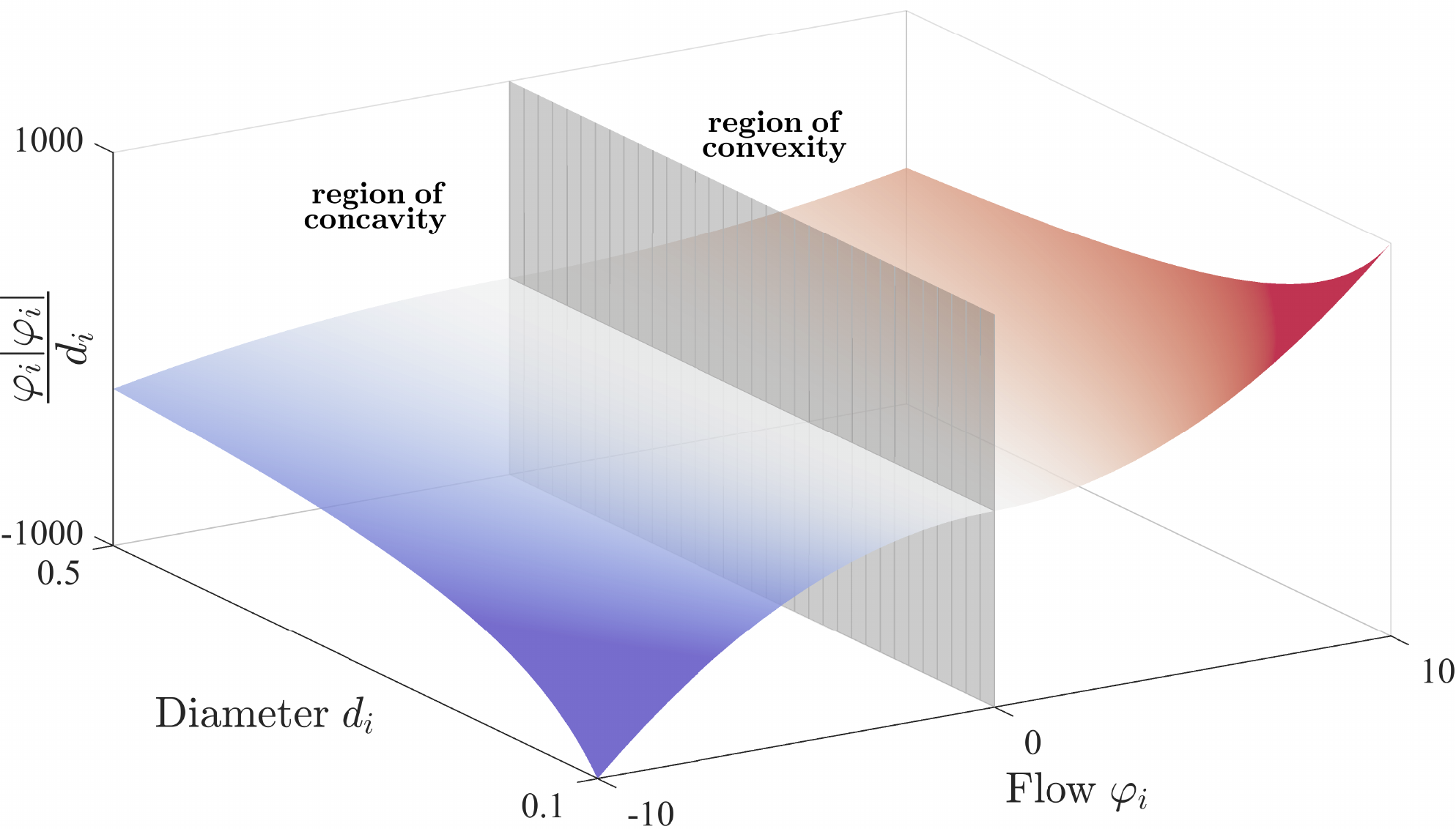}
    \caption{The concave region of the function $\varphi_{i}|\varphi_{i}|/d_{i}$ is depicted on the left (for $\varphi_{i}\le0$) in blue, and the convex region is depicted on the right (for $\varphi_{i}\ge0$) in red.}
    \label{fig:convex_concave}
\end{figure}
\section{Modeling diameter-dependent Weymouth equations}\label{app:Weymputh_explantion_details}
The pipeline friction $\omega_i$ from \eqref{OGF:weymouth} incorporates the fixed pipeline length, diameter, rugosity, as well as  gas temperature, compressibility and density relative to air~\cite{de2000gas}. There are a spectrum of different methods available for modeling pipeline friction coefficients, but in this paper, we use the simplifying assumption that $\omega_i$ is linearly proportional to diameter of the line $d_i$, as in~\cite{Sundar:2019} (i.e., the Darcy—Wiesbach friction factor is assumed constant, regardless of the pipeline's diameter). From \eqref{expantion:Weymouth}, we then seek to use ICNNs to build a surrogate model which mimics the rational expression
\begin{align}\label{f_rational}
f(\varphi_{i},d_{i})=\frac{\varphi_{i}|\varphi_{i}|}{d_{i}}.
\end{align}
Despite its nonlinearity, \eqref{f_rational} can be expressed as the sum of one convex function ($f_{{\rm +}}$) and one concave function ($f_{{\rm +}}$) across its domain of practical usage: $f(\varphi_{i},d_{i})=f_{{\rm +}}(\varphi_{i},d_{i})+f_{-}(\varphi_{i},d_{i})$. Figure~\ref{fig:convex_concave} depicts the convex and concave regions of $f(\varphi_{i},d_{i})$.

\section{MIQP relaxation of the operational planning problem}\label{app:MIQP_relaxation}

The quadratic programming relaxation of the static Weymouth equation \eqref{OGF:weymouth} is
\begin{align}\label{relax_aux_1}
    &\varphi\circ|\varphi|=\text{diag}[\omega]A^{\top}\pi\overset{\text{relax.}}{\Longrightarrow}
    \left\{
    \begin{matrix*}[l]
        \pi_{i} -\pi_{j} \geqslant \frac{1}{\omega_{l}}\varphi_{l}^{2} & \text{if}\;\varphi\geqslant0\\
        \pi_{j} -\pi_{i} \geqslant \frac{1}{\omega_{l}}\varphi_{l}^{2} &\text{if}\;\varphi\leqslant0
    \end{matrix*}
    \right.\;\forall l=(i,j)\in1,\dots,\ell,
\end{align}
which distinguishes between two cases of either positive or negative gas flow in every line $l$ with sending and receiving ends denoted by $i$ and $j$, respectively. Using a binary variable $x_{l}$, this relaxation can be written as 
\begin{align}\label{relax_aux_2}
    &(2x_{l} - 1)\pi_{i} + (1 - 2x_{l})\pi_{j} \geqslant  \frac{1}{\omega_{l}}\varphi_{l}^{2}, \quad \forall \ell=(i,j)\in1,\dots,\ell,
\end{align}
such that, when $x_{l}=1$, we have $\varphi_{l}\geqslant0$, and when $x_{l}=0$, we have $\varphi_{l}\leqslant0$. The bilinear terms $x_{l}\pi_{i}$ and $x_{l}\pi_{j}$ can be handled using the Big-M method. Using an auxiliary variable $z_{l i}=x_{l}\pi_{i}$, the first bilinear term can be restated as follows:
\begin{subequations}
\begin{align}
    x_{l}\underline{\pi}_{i} \leqslant  &\; z_{li} \leqslant x_{l}\overline{\pi}_{i}, \\
    \pi_{i}+(x_{l}-1)\overline{\pi}_{i} \leqslant &\;z_{li} \leqslant \pi_{i}+(x_{l}-1)\underline{\pi}_{i}, 
\end{align}
\end{subequations}
where $\underline{\pi}_{i}$ and $\overline{\pi}_{i}$ respectively denote the lower and upper pressure limits at node $i$. When $x_{l}=1$, $z_{li}=\pi_{i}$, and when $x_{l}=0$, $z_{li}=0$. Then, the MIQP  relaxation of the Weymouth equation is
\begin{subequations}\label{relax_aux_3}
\begin{align}
    &2z_{li} - 2z_{lj} - \pi_{i} + \pi_{j} \geqslant  \tfrac{1}{\omega_{l}}\varphi_{l}^{2}, \quad \forall l=(i,j)\in1,\dots,\ell,\\
    &x_{l}\underline{\pi}_{i} \leqslant   z_{li} \leqslant x_{l}\overline{\pi}_{i}, \quad\forall i=1,\dots,n,\;\forall l\in1,\dots,\ell\\
    &\pi_{i}+(x_{l}-1)\overline{\pi}_{i} \leqslant z_{li} \leqslant \pi_{i}+(x_{l}-1)\underline{\pi}_{i},\quad\forall i\in1,\dots,n,\;\forall l\in1,\dots,\ell. 
\end{align}
\end{subequations}
Substituting the Weymouth equation \eqref{OGF:weymouth} with equations \eqref{relax_aux_3} yields the relaxed planning problem. 

\end{document}